\documentclass[10pt,journal,twocolumn]{IEEEtran}

\usepackage[english]{babel}
\usepackage[T1]{fontenc}
\usepackage{url}

\usepackage[cmex10]{amsmath}
\usepackage{amsthm,amsfonts,amssymb}

\usepackage{graphicx}
\usepackage[caption = false]{subfig}
\usepackage{color}

\usepackage{colortbl}
\usepackage{url}
\usepackage{booktabs}
\usepackage{epsfig}
\usepackage{pstricks}
\usepackage{pst-node}
\usepackage{pst-grad}
\usepackage{pst-plot}
\usepackage{pstricks-add}
\usepackage{ifthen}
\usepackage{algorithm}
\usepackage{algpseudocode}
\usepackage{float}
\usepackage{fp}
\usepackage{cite}
\usepackage{multirow}
\usepackage{hhline}

\newfloat{algorithm}{t}{lop}

\newboolean{draft}
\newcommand{\isdraft}[2]{\ifthenelse{\boolean{draft}}{#1}{#2}}

\isdraft{\usepackage{setspace}}{}                              % DRAFT
\isdraft{\usepackage[footnotesize]{caption}}{}                 % DRAFT
\isdraft{\usepackage{paralist}}{}                              % DRAFT

\theoremstyle{plain}

\definecolor{red}{RGB}{153,0,0}
\definecolor{green}{RGB}{0,153,0}			
\definecolor{blue}{RGB}{0,0,153}
\definecolor{darkred}{RGB}{90,0,0}
\definecolor{darkgreen}{RGB}{0,90,0}
\definecolor{darkblue}{RGB}{0,0,90}

\newcommand*{\tran}{^{\mkern-1.5mu\mathsf{T}}}

\isdraft{}{}

\title{Deep Coupled-Representation Learning for Sparse Linear Inverse Problems with Side Information }
\author{Evaggelia~Tsiligianni
        and Nikos~Deligiannis% <-this % stops a space
\thanks{Both authors are with the Department
of Electronics and Informatics, Vrije Universiteit Brussel, Brussels, Belgium, 
and with imec, Kapeldreef 75, B-3001, Leuven, Belgium. email: \texttt{\{etsiligi, ndeligia\}@etrovub.be}.}% <-this % stops a space
}

\begin{document}

\maketitle

\begin{abstract}
In linear inverse problems, the goal is to recover a target signal 
from undersampled, incomplete or noisy linear measurements.
Typically, the recovery relies on complex numerical optimization methods;
recent approaches perform an unfolding of a numerical algorithm into a neural network form,
resulting in a substantial reduction of the computational complexity.
In this paper, we consider the recovery of a target signal 
with the aid of a correlated signal, the so-called \textit{side information (SI)},
and propose a deep unfolding model that incorporates SI.
The proposed model is used to learn coupled representations of correlated signals from different modalities,
enabling the recovery of multimodal data at a low computational cost.
As such, our work introduces the first deep unfolding method with SI, which actually comes from a different modality.
We apply our model to reconstruct near-infrared images from undersampled measurements
given RGB images as SI.
Experimental results demonstrate the superior performance of the proposed framework 
against single-modal deep learning methods that do not use SI, 
multimodal deep learning designs, and optimization algorithms.
\end{abstract}

%\begin{keywords}
%Inverse problems, coupled representations, representation learning, designing deep neural networks.
%\end{keywords}

%============================================%
\section{Introduction}
\label{sec:intro}
%============================================%
Linear inverse problems arise in various signal processing domains
such as computational imaging, remote sensing, seismology and astronomy, to name a few. 
These problems can be expressed by a linear equation of the form:
\begin{equation}
\label{eq:system}
y = \Phi x + e,
\end{equation}
where $x \in \mathbb{R}^n$ is the unknown signal,
$\Phi \in \mathbb{R}^{m \times n}$, $m \ll n$, is a linear operator,
and $y \in \mathbb{R}^m$ denotes the observations contaminated with noise $e \in \mathbb{R}^m$. 
Sparsity is commonly used for the regularization of ill-posed inverse problems,
leading to the so-called sparse approximation problem~\cite{tropp2010computational}. 
Compressed sensing (CS)~\cite{donoho2006compressed} deals with 
the sparse recovery of linearly subsampled signals and falls in this category.

In several applications, besides the observations of the target signal,
additional information from correlated signals is often 
available~\cite{Zhang2008, weizman2015compressed, mota2015dynamic, vaswani2010modified, Ma2019AMPSI, song2016coupled, deligiannis2017multi, song2017multimodal}.
In multimodal applications,  combining information from multiple signals calls for methods
that allow coupled signal representations, capturing the similarities between correlated data.
To this end, coupled dictionary learning is a popular approach~\cite{song2016coupled, deligiannis2017multi, song2017multimodal};
however,  dictionary learning methods employ overcomplete dictionaries,
resulting in computationally expensive sparse approximation problems.

Deep learning has gained a lot of momentum in solving inverse problems, 
often surpassing the performance of 
analytical approaches~\cite{lucas2018using, nguyen2017deep, mousavi2017learning}. 
Nevertheless, neural networks have a complex structure and appear as ``black boxes''; 
thus, understanding what the model has learned is an active research topic.
Among the efforts trying to bridge the gap between analytical methods and deep learning
is the work presented in~\cite{LeCun_LISTA}, which introduced the idea of 
unfolding a numerical algorithm for sparse approximation into a neural network form.
Several unfolding approaches~\cite{xin2016maximal, hershey2014deep, borgerding2017amp} followed that of~\cite{LeCun_LISTA}.
Although the primary motivation for deploying deep learning in inverse problems
concerns the reduction of the computational complexity,
unfolding offers another significant benefit:
the model architecture allows a better insight in the inference procedure 
and enables the theoretical study of the network 
using results from sparse modelling~\cite{Chen2018convergence, giryes2018tradeoffs, papyan2017convolutional, xin2016maximal}.

In this paper, we propose a deep unfolding model for the recovery of a signal 
with the aid of a correlated signal, the \textit{side information (SI)}.
To the best of our knowledge, this is the first work in deep unfolding that incorporates SI.
Our contribution is as follows:
(i)~Inspired by~\cite{LeCun_LISTA}, we design a deep neural network that unfolds a proximal algorithm for sparse approximation with SI; we coin our model Learned Side Information Thresholding Algorithm (LeSITA).
(ii)~We use LeSITA in an autoencoder fashion to learn coupled representations of correlated signals from different modalities.
(iii)~We design a LeSITA-based reconstruction operator that utilizes learned SI provided by the autoencoder
to enhance signal recovery. 
 
%Our method is applied to reconstruct near-infrared images from CS measurements with the aid of RGB images.
We test our method in an example application, namely, multimodal reconstruction from CS measurements.
Other inverse problems of the form~\eqref{eq:system}
such as image super-resolution~\cite{liu2016, song2016coupled} or image denoising~\cite{Metzler2016denoising}
can benefit from the proposed approach.
We compare our method with existing single-modal deep learning methods that do not use SI, 
multimodal deep learning designs, and optimization algorithms, showing its superior performance.

The paper is organized as follows.
Section~\ref{sec:background} provides the necessary background and reviews related work.
The proposed framework is presented in Section~\ref{sec:proposed}, 
followed by experimental results in Section~\ref{sec:experiments}.
Conclusions are drawn in Section~\ref{sec:conclusions}.

%============================================%
\section{Background and Related Work}
\label{sec:background}
%============================================%
%
\begin{figure}
\centering
\subfloat[]{\includegraphics[scale=0.55]{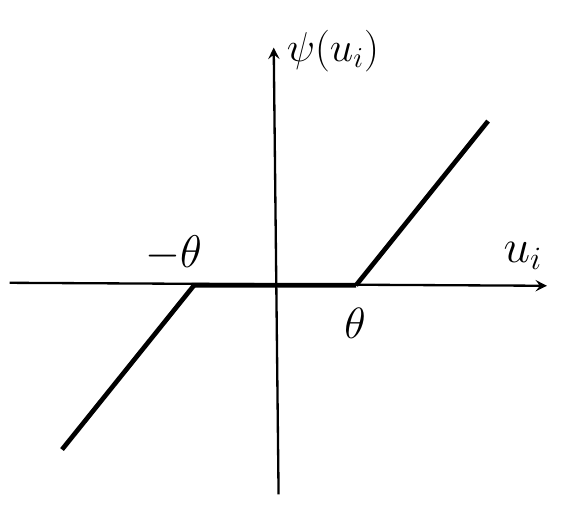}}
\hspace{0.7cm}
\subfloat[]{\includegraphics[scale=0.55]{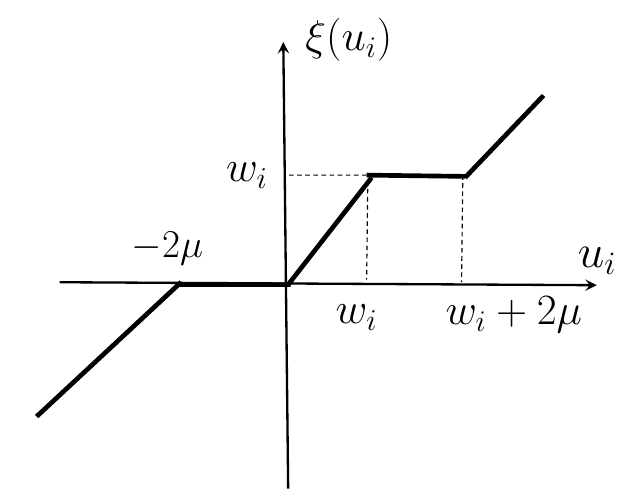}}
 \caption{Graphical representation of the proximal operators of (a)\ ISTA  and (b)\ SITA
 (for non-negative SI $w_i \geq 0$, $i=1, \dots , k$). $\theta, \mu$ are positive parameters.}   
 \label{fig:proximals}
\end{figure}
%
%%--------------------------------------------------------------------------%
%\subsection{Sparse Approximation with Proximal Methods}
%%--------------------------------------------------------------------------%
A common approach for solving problems of the form~\eqref{eq:system} 
with sparsity constraints is convex optimization~\cite{chen2001atomic}.
Let us assume that the unknown $x \in \mathbb{R}^{n}$ has a sparse representation $\alpha \in \mathbb{R}^{k}$ 
with respect to a dictionary $D_x \in \mathbb{R}^{n \times k}$, $n \leq k$,
that is, $x = D_x \alpha$. 
Then, \eqref{eq:system} takes the form
\begin{equation}
\label{eq:system_sparse}
y = \Phi D_x \alpha + e,
\end{equation}
and a solution can be obtained via the formulation of the $\ell_1$ minimization problem:
\begin{equation}
\min_{\alpha} \frac{1}{2}\| \Phi D_x \alpha-y\|_2^2  + \lambda \|\alpha\|_1,
\label{eq:probleml1}
\end{equation}
where $\|\cdot\|_1$ denotes the $\ell_1$-norm ($\|\alpha\|_1 = \sum_{i=1}^{n} |\alpha_i|$),
which promotes sparse solutions and $\lambda$ is a regularization parameter.

Numerical methods~\cite{tropp2010computational} proposed to solve~\eqref{eq:probleml1}
include pivoting algorithms, interior-point methods,  gradient based methods
and message passing algorithms (AMP)~\cite{rangan2011generalized}.
Among gradient based methods, proximal methods are tailored to optimize an objective of the form
\begin{equation}
\label{eq:problem_basic}
\min_{\alpha} f(\alpha) + \lambda g(\alpha),
\end{equation}
where $f:\mathbb{R}^n \rightarrow \mathbb{R}$ is a convex differentiable function with a Lipschitz-continuous gradient, 
and $g:\mathbb{R}^n \rightarrow \mathbb{R}$ is convex and possibly nonsmooth~\cite{combettes2011proximal}, \cite{Bach2012}.
Their main step involves the proximal operator, defined for a function $g$ according to
\begin{equation}
\label{proxdef}
\text{prox}_{\theta g} (u) = \arg\min_{v}  \big\{ \frac{1}{2} \|v-u\|_2^2 + \theta g(v) \big\},
\end{equation}
with $\theta = \frac{\lambda}{L}$ and $L>0$ an upper bound on the Lipschitz constant of $\nabla f$. 
A popular proximal algorithm is the Iterative Soft Thresholding Algorithm (ISTA)~\cite{daubechies2004iterative, combettes2005}.
Let us set $F := \Phi D_x$,  $F  \in \mathbb{R}^{m \times k}$ in~\eqref{eq:probleml1}. At the $t$-th iteration ISTA computes:
\begin{equation}
\label{eq:ista}
\alpha^{t} = \psi_{\theta} (\alpha^{t-1} - \frac{1}{L} F\tran (F \alpha^{t-1} -y) ), \quad \alpha^0 = 0,
\end{equation}
where $\psi_{\theta }$ denotes the proximal operator~[Figure~\ref{fig:proximals}(a)]
expressed by the component-wise shrinkage function:
\begin{equation}
\label{eq:proxl1}
\psi_{\theta}(u_i) = \text{sign}(u_i)(|u_i| - \theta)_+, \quad  i=1, \dots , k,
\end{equation}
with $u_+=\max\{u, 0\}$.

In order to account for the high computational cost of numerical algorithms,
Gregor and LeCun~\cite{LeCun_LISTA} unfolded ISTA into a neural network referred to as LISTA.
% LISTA
Specifically, by setting $S=I- \frac{1}{L} F\tran F$, $W=\frac{1}{L} F\tran $, 
\eqref{eq:ista} results in
\begin{equation}
\label{eq:lista2}
\alpha^{t} = \psi_{\theta} \big(S \alpha^{t-1} +Wy \big).
\end{equation}
Considering a correspondence of every iteration with a neural network layer,
a number of iterations of (\ref{eq:lista2}) can be implemented by a recurrent or feed forward neural network;
$S$, $W$ and $\theta$ are learnable parameters, 
and the proximal operator (\ref{eq:proxl1}) acts as a nonlinear activation function.
A fixed depth network allows the computation of sparse codes in a fixed amount of time.
Similar unfolding methods were proposed in~\cite{xin2016maximal, hershey2014deep, borgerding2017amp}.
%============================================%
\section{Proposed Framework}
\label{sec:proposed}
%============================================%
In this paper, we consider that, besides the observations of the target signal, 
we also have access to SI, that is, a signal $z$ correlated to the unknown $x$.
We assume that $x \in \mathbb{R}^{n}$ and $z \in  \mathbb{R}^{d}$ 
have similar sparse representations $\alpha \in \mathbb{R}^{k}$, $w \in \mathbb{R}^{k}$, 
under dictionaries $D_x \in \mathbb{R}^{n \times k}$, $D_z \in \mathbb{R}^{d \times k}$, $n \leq k$, $d \leq k$, respectively.
Specifically, we assume that  $\alpha$ and $w$ are similar by means of the $\ell_1$ norm, that is, $\|\alpha-w\|_1$ is small.
The condition holds for representations with partially common support and a number of similar nonzero coefficients;
we refer to them as \textit{coupled sparse representations}.
Then, $\alpha$ can be obtained from the $\ell_1$-$\ell_1$ minimization problem
\begin{equation}
\min_{\alpha} \frac{1}{2}\| \Phi D_x \alpha-y\|_2^2  + \lambda (\|\alpha\|_1  +  \|\alpha-w\|_1).
\label{eq:probleml1-l1}
\end{equation}
\eqref{eq:probleml1-l1} has been theoretically studied in~\cite{nikosIT}
and has been employed for the recovery of sequential signals in~\cite{Zhang2008, weizman2015compressed, mota2015dynamic}.

We can easily obtain coupled sparse representations of sequential signals that change slowly
using the same sparsifying dictionary~\cite{Zhang2008, weizman2015compressed, mota2015dynamic}.
However, this is not the case in most multimodal applications,
where, typically, finding coupled sparse representations involves dictionary learning 
and complex optimization methods~\cite{song2016coupled, deligiannis2017multi, song2017multimodal}.
In this work, we propose an efficient approach based on a novel multimodal deep unfolding model.
The model is employed for learning coupled representations of the target signal and the SI (Section~\ref{sec:lesita_ae}),
and for reconstruction with SI (Section~\ref{sec:lesita_rec}).
Our approach is inspired by a proximal algorithm for the solution of~\eqref{eq:probleml1-l1}.

%--------------------------------------------------------------------------------------------------%
\subsection{ Sparse Approximation with SI via Deep Unfolding}
\label{sec:lesita}
%--------------------------------------------------------------------------------------------------%
Problem~\eqref{eq:probleml1-l1} is of the form \eqref{eq:problem_basic} with
$f(\alpha) = \frac{1}{2}\|F \alpha-y\|_2^2$, $F:= \Phi D_x$, $F  \in \mathbb{R}^{m \times k}$,
and $g( \alpha ) =  \| \alpha\|_1 + \|\alpha-w\|_1$.
The proximal operator for $g$ is defined by
\begin{equation}
\label{eq:proxLeSITA}
\xi_{\mu} (u) = \arg\min_{v}  \big\{ \frac{1}{2} \|v-u\|_2^2 + \mu (\|v\|_1 + \|v-w\|_1) \big\},
\end{equation}
where $ \mu = \frac{\lambda}{L}$, and $L>0$ is an upper bound on the Lipschitz constant of $ \nabla f$.
All terms in (\ref{eq:proxLeSITA}) are separable, thus,  we can easily show that (see Appendix):
\begin{enumerate}
\item{For $w_i\geq 0$,  $i=1, \dots , k$:
\begin{equation}
\label{eq:prox_pos}
\xi_{\mu}(u_i) = 
\left\{\begin{alignedat}{2}
&u_i+2\mu,	\quad	&&u_i<-2\mu, \\
&0,			\quad	&-2\mu \leq &u_i \leq 0, \\
&u_i,			\quad	&0 <&u_i<w_i, \\
&w_i,		\quad	&w_i \leq &u_i \leq w_i + 2\mu, \\
&u_i-2\mu	,	\quad	&&u_i> w_i+ 2\mu.
\end{alignedat} \right.
\end{equation}
}
\item{For $w_i < 0$, $i=1, \dots , k$:
\begin{equation}
\label{eq:prox_neg}
\xi_{\mu}(u_i) = 
\left\{\begin{alignedat}{2}
&u_i+2\mu,	\quad	&&u_i<w_i-2\mu, \\
&w_i,		\quad	&w_i-2\mu \leq &u_i \leq w_i, \\
&u_i,			\quad	&w_i <&u_i<0, \\
&0,			\quad	&0 \leq &u_i \leq   2\mu, \\
&u_i-2\mu	,	\quad	&&u_i>  2\mu.
\end{alignedat} \right.
\end{equation}
}
\end{enumerate}
Figure~\ref{fig:proximals}(b) depicts the graphical representation of the proximal operator given by~\eqref{eq:prox_pos}.
With $\nabla f(\alpha) =  F\tran (F \alpha -y)$, a proximal method for ($\ref{eq:probleml1-l1}$) takes the form
\begin{equation}
\label{eq:SITA}
\alpha^{t} = \xi_{\mu} (\alpha^{t-1} - \frac{1}{L} F\tran (F \alpha^{t-1} -y) ), \quad \alpha^0 = 0.
\end{equation}
We coin \eqref{eq:SITA} Side-Information-driven iterative soft Thresholding Algorithm  (SITA).
%%----------------------------------------------------------------------------------------------%
%\subsection{Learned SITA}
%\label{sec:lesita}
%%----------------------------------------------------------------------------------------------%

\begin{figure}[t]
\centering
\includegraphics[scale=0.5]{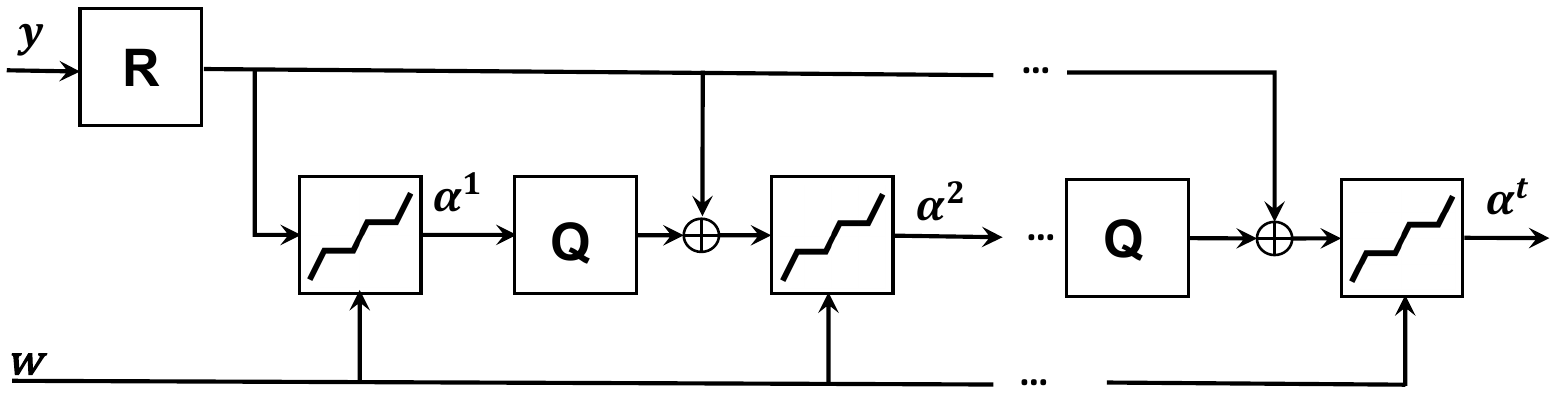}
\caption{LeSITA: Unfolding SITA~\eqref{eq:SITA} to a neural network form~\eqref{eq:LESITA1}.}
\label{fig:LeSITAunfolded}
\end{figure}

We unfold SITA to a neural network form,
by settting $Q=I- \frac{1}{L} F\tran F$, $R=\frac{1}{L} F\tran$.
Then~\eqref{eq:SITA} results in
\begin{equation}
\label{eq:LESITA1}
\alpha^{t} = \xi_{\mu} \big(Q \alpha^{t-1} +Ry \big).
\end{equation}
\eqref{eq:LESITA1} has a similar expression to LISTA~\eqref{eq:lista2};
however, the two algorithms involve different proximal operators (Figure~\ref{fig:proximals}).
A fixed number of iterations of (\ref{eq:LESITA1}) can be implemented 
by a recurrent or feed forward neural network,
with the proximal operator given by (\ref{eq:prox_pos}), (\ref{eq:prox_neg}) 
employed as a nonlinear activation function, which integrates the SI;
$Q$, $R$ and $\mu$ are learnable parameters.
The network architecture is depicted in Figure~\ref{fig:LeSITAunfolded}.

We can train the neural network using $J$ pairs of sparse codes $\{\alpha_{(j)}, w_{(j)}\}_{j=1}^{J}$ 
corresponding to $J$ pairs of correlated signals $\{x_{(j)}, z_{(j)}\}_{j=1}^{J}$,  
and a loss function of the form:
\begin{equation}
\label{eq:LESITAloss}
\mathcal{L} = \sum_{j=1}^{J}\|\alpha_{(j)}-\hat{\alpha}_{(j)}\|_2^2,
\end{equation}
where $\hat{\alpha}_{(j)}$ is the output estimation.
The learning results in a fast sparse approximation operator 
that directly maps the input observation vector $y$ to a sparse code $\alpha$ with the aid of the SI $w$.
We coin this operator Learned Side Information Thresholding Algorithm (LeSITA).
Being based on an optimization method, LeSITA can be theoretically analyzed
(see~\cite{Ma2019AMPSI, Chen2018convergence, giryes2018tradeoffs, papyan2017convolutional, xin2016maximal}).
We leave this analysis for future work.%
%----------------------------------------------------------------------------------------------%
\subsection{LeSITA Autoencoder for Coupled Representations}
\label{sec:lesita_ae}
\begin{figure}
\centering
\includegraphics[scale=0.45]{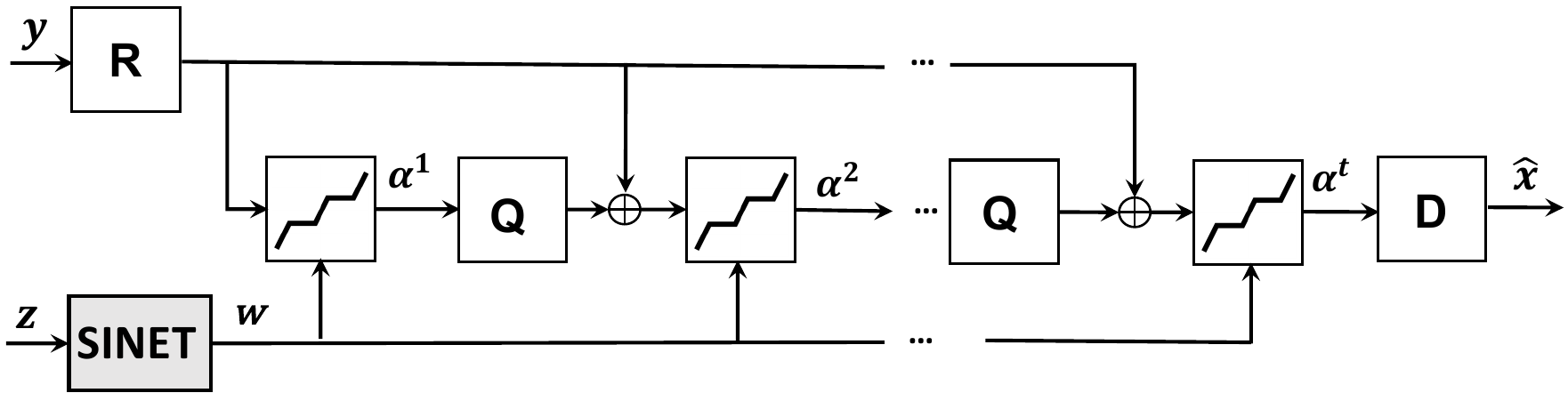}
\caption{Use of LeSITA for signal representation or reconstruction with SI. 
The main branch comprises a LeSITA encoder and a linear decoder;
the input is either the signal $x$ (Section~\ref{sec:lesita_ae}) or the observations $y$ (Section~\ref{sec:lesita_rec}).
The SI branch (SINET) performs transformation of the SI.
The transformed SI $w$ is used to guide LeSITA
to produce a representation $\alpha$ of the target signal that improves reconstruction.}
\label{fig:LeSITArecGeneral}
\end{figure}

%----------------------------------------------------------------------------------------------%
%Instead of training using sparse codes,
%we can use LeSITA in an autoencoder fashion
%to build a learning framework for coupled representations of $x$, $z$.
Instead of training using sparse codes,
we can use LeSITA in an autoencoder fashion
to learn coupled representations of $x$, $z$.
By setting $\Phi$  equal to the identity matrix, 
\eqref{eq:probleml1-l1} reduces to a sparse representation problem with SI.
Then,~\eqref{eq:LESITA1} can compute a representation of $x$
according to $\alpha^{t} = \xi_{\mu} \big(Q \alpha^{t-1} +R x \big)$. 
The proposed autoencoder is depicted in Figure~\ref{fig:LeSITArecGeneral}.
The main branch accepts as input the target signal $x$  ($y = x$). 
The core component is a LeSITA encoder,
followed by a linear decoder performing reconstruction, i.e., $\hat{x} = D \alpha$;
$D \in \mathbb{R}^{n \times k}$ is a trainable dictionary ( $D$ is not tied to any other weight).
A second branch referred to as SINET acts as an SI encoder,
performing a (possibly) nonlinear transformation of the SI. 
We employ LISTA~\eqref{eq:lista2} to incorporate sparse priors in the transformation,
obtaining $w^{t} = \psi_{\theta} \big(S w^{t-1} +W z \big)$, $w^0 = 0$;
$\psi_{\theta}$ is given by \eqref{eq:proxl1}, and $S$, $W$ and $\theta$ are learnable parameters.
The number of layers of LISTA and LeSITA may differ.

We use $J$ pairs of correlated signals $\{x_{(j)}, z_{(j)}\}_{j=1}^{J}$ to train our autoencoder,
and an objective function of the form:
\begin{equation}
\label{eq:LESITAaeloss}
\mathcal{L} = \lambda_1\mathcal{L}_{1}+ \lambda_2\mathcal{L}_{2},
\end{equation}
where $\mathcal{L}_{1}$ is the reconstruction loss,
$\mathcal{L}_{2}$ is a constraint on the latent representations,
and $\lambda_1$, $\lambda_2$ are appropriate weights. 
We use the $\ell_2$ norm as reconstruction loss,
i.e., $\mathcal{L}_{1} = \sum_{j=1}^{J}\|x_{(j)}-\hat{x}_{(j)}\|_2^2$,
where $x_{(j)}$ is the $j$-th sample of the target signal and $\hat{x}_{(j)}$ is the respective output estimation.
We set $\mathcal{L}_{2} = \sum_{j=1}^{J}\|\alpha_{(j)}-w_{(j)}\|_1$
to promote coupled latent representations capturing the correlation between $x_{(j)}$ and $z_{(j)}$.
\begin{table}
\centering
\caption{Sparse approximation results (NMSE in dB).}
\label{table: compare_sparse}
\setlength\tabcolsep{1.5pt}
\begin{tabular}{ | c || c || c | c | c | c || c | c |}
\hline 
%    	&\multirow{2}{*}{\cite{LeCun_LISTA}}		&\multicolumn{4}{c|} {LeSITA}  \\
	&{\cite{LeCun_LISTA}}		&\multicolumn{4}{c||} {LeSITA} &\multicolumn{2}{c|} {SITA} \\
\hhline{|-||-|-|-|-|-||-|-|}
similarity    & -- 				&$\rho = 25$ 	&$\rho = 20$  	&$\rho = 15$  	&$\rho = 10$   		&$\rho = 25$  	&$\rho = 15$\\
 \hline \hline 
$T=3$    	&$-15.64$  		&$-21.10$ 	&$-18.05$		&$-15.54$  	&$-13.45$   		&$-2.25$ 		&$-2.22$ \\
\hline
$T=5$   	&$-21.92$  		&$-27.97$ 	&$-24.67$		&$-21.68$  	&$-18.85$   		&$-2.70$		&$-2.65$ \\
\hline
$T=7$ 	&$-26.95$  		&$-33.54$ 	&$-30.06$		&$-26.84$  	&$-23.87$   		&$-3.00$		&$-2.94$ \\
\hline
\end{tabular}
\end{table}
%%%
%
\begin{table*}[ht]
\centering
\caption{Reconstruction results on NIR images (PSNR in dB).}
\label{table:compare_rec}
\setlength\tabcolsep{5pt}
\begin{tabular}{ | c || c | c || c | c || c | c || c | c || c | c |c | c | }
\hline 
&\multicolumn{6}{c||}{Single-modal methods} &\multicolumn{6}{c|}{Multi-modal methods}\\
\hline 
&\multicolumn{2}{c||}{\textbf{LISTA~\cite{LeCun_LISTA}}} 	&\multicolumn{2}{c||}{\textbf{LAMP~\cite{borgerding2017amp}}}  &\multicolumn{2}{c||} {\textbf{DL~\cite{nguyen2017deep}}}  &\multicolumn{2}{c||} {\textbf{Multimodal DL}}  &\multicolumn{2}{c|} {\textbf{LeSITA ($\mathcal{L}_{2}^{\text{B}}$)}} &\multicolumn{2}{c|} {\textbf{LeSITA ($\mathcal{L}_{2}^{\text{A}}$)}}   \\  
\hline 
%						% LISTA		% LISTA		% LAMP		% LAMP			%DUC		%DUC		%MULTIMODAL			% SPARSE LESITA		%PROPOSED
\textbf{CS ratio}   			&$0.5$  		&$0.25$ 		&$0.5$  		&$0.25$  			&$0.5$  		&$0.25$ 		&$0.5$  		&$0.25$ 		&$0.5$&	$0.25$ 		&$0.5$  		&$0.25$\\
\hline \hline
\textbf{country (0070)}  		&$45.64$  	&$38.16$ 		&$44.98$   	&$36.76$  		&$34.04$ 		&$32.84$  	&$40.69$ 		&$38.52$ 		&$41.99$	&$35.48$ 		&$\bf46.34$ 		&$\bf39.80$\\
\hline
\textbf{field (0058)}	  		&$40.01$  	&$33.94$  	&$39.87$ 		&$33.17$  		&$31.02$ 		&$30.66$ 		&$36.13$ 		&$34.35$		&$37.74$	&$32.47$ 		&$\bf40.61$ 		&$\bf35.65$\\
\hline
\textbf{forest (0058)}	 		&$37.82$  	&$31.69$ 		&$37.69$ 		&$30.80$  		&$28.50$ 		&$28.28$ 		&$33.91$ 		&$32.05$		&$35.49$	&$29.97$ 		&$\bf38.54$ 		&$\bf34.03$\\
\hline 
\textbf{indoor (0056)  }		&$37.18$  	&$32.05$  	&$37.05$ 		&$31.17$  		&$29.08$ 		&$28.85$ 		&$33.84$ 		&$32.42$		&$35.19$	&$30.72$ 		&$\bf37.90$ 		&$\bf34.93$\\
\hline
\textbf{mountain (0055)} 		&$54.33$  	&$53.53$ 		&$56.13$ 		&$51.96$  		&$51.12$ 		&$45.26$ 		&$54.62$ 		&$53.20$		&$55.45$	&$52.25$ 		&$\bf56.79$ 		&$\bf53.74$\\
\hline 
\textbf{oldbuilding (0103)}  	&$49.04$  	&$41.51$  	&$48.17$ 		&$39.00$  		&$36.05$ 		&$34.07$ 		&$44.39$ 		&$41.82$		&$44.47$	&$40.33$ 		&$\bf50.69$ 		&$\bf44.20$\\
\hline 
\textbf{street (0057)}  		&$37.61$  	&$33.10$  	&$36.09$ 		&$31.45$  		&$31.01$ 		&$29.79$ 		&$34.30$ 		&$33.22$		&$36.46$	&$32.38$ 		&$\bf38.13$ 		&$\bf35.13$\\
\hline 
\textbf{urban (0102)} 		&$\bf40.00$  	&$32.88$  	&$38.78$ 		&$31.95$  		&$29.55$ 		&$29.22$ 		&$35.11$ 		&$33.28$		&$36.47$	&$31.69$ 		&$39.69$	 		&$\bf35.05$\\
\hline 
\textbf{water (0083)}  		&$46.79$  	&$42.50$  	&$47.10$ 		&$41.25$  		&$38.43$ 		&$35.47$ 		&$44.21$ 		&$42.69$		&$45.24$	&$41.35$ 		&$\bf47.69$ 		&$\bf43.58$\\
\hline \hline
\textbf{Average}	  		&$43.16$  	&$37.71$  	&$42.87$ 		&$36.39$  		&$34.31$ 		&$32.72$ 		&$39.69$ 		&$37.95$ 		&$40.94$	&$36.29$ 		&$\bf44.04$ 		&$\bf39.57$\\
\hline 
\end{tabular}
\end{table*}
%%%
%----------------------------------------------------------------------------------------------%
\subsection{LeSITA for Reconstruction with SI}
\label{sec:lesita_rec}
%----------------------------------------------------------------------------------------------%
We propose a reconstruction operator that effectively utilizes SI for signal recovery,
following the architecture of 
Figure~\ref{fig:LeSITArecGeneral}.
In the main branch, a LeSITA encoder
computes a latent representation $\alpha$
of the observation vector $y$ obtained from~\eqref{eq:system}, %(corresponding to the unknown signal $x$)
according to~\eqref{eq:LESITA1}.
A linear decoder performs reconstruction of the unknown signal, 
i.e., $\hat{x} = D \alpha$;
$D \in \mathbb{R}^{n \times k}$ is a learnable dictionary.
The role of the SINET branch is to enhance the encoding process
by providing LeSITA with prior knowledge.
In this task, the SINET is realized by a LISTA encoder,
the weights of which are initialized with the SINET weights of the trained autoencoder~(Sec.~\ref{sec:lesita_ae}).
In this way, the LeSITA autoencoder is used to provide coupled sparse representations.
%We leverage the proposed LeSITA autoencoder 
%to obtain coupled representations of the target and the SI signals,
%by realizing SINET with a LISTA encoder, which is initialized with weights 
%from an autoencoder learned to reconstruct $x$ (Sec.~\ref{sec:lesita_ae}).
The proposed model is trained using the $\ell_2$ loss function, $\mathcal{L} = \sum_{j=1}^{J}\|x_{(j)}-\hat{x}_{(j)}\|_2^2$,
with $x_{(j)}$ the $j$-th sample of the target signal 
and $\hat{x}_{(j)}$ the respective model estimation.
%============================================%
\section{Experimental results}
\label{sec:experiments}
%============================================%
A first set of experiments concerns the performance of the proposed LeSITA model~\eqref{eq:LESITA1} 
in sparse approximation using synthetic data. 
We generate $J=500$K pairs of sparse signals $\{\alpha_{(j)}, w_{(j)}\}_{j=1}^{J}$ of length $k = 256$
with $s=25$ nonzero coefficients drawn from a standard normal distribution.
The sparsity level is kept fixed but the signals have varying support.
The SI is generated such that
$\alpha_{(j)}$ and $w_{(j)}$ share the same support $\mathcal{I}_{(j)}$ in a number of positions $\rho \leq s$,
that is, $\mathcal{I}_{(j)} = \{i: w_{(j)}[i]\neq 0, \alpha_{(j)}[i] \neq 0\}$, $|\mathcal{I}_{(j)}| = \rho$,
with $\alpha_{(j)}[i]$, $w_{(j)}[i]$ denoting the $i$-th coefficient of the respective signals.
For $i \in \mathcal{I}_{(j)}$, we obtain $w_{(j)}[i] = \lvert \kappa \lvert \alpha_{(j)}[i]$, 
where $\kappa$ is drawn from a normal distribution;
therefore, for $i \in \mathcal{I}_{(j)}$, 
the coefficients of $\alpha_{(j)}$ and $w_{(j)}$ are of the same sign;
the rest are drawn from a standard normal distribution.
We vary the values of $\rho$, i.e., $\rho=\{25, 20, 15, 10\}$,  
to obtain different levels of similarity between $\alpha$ and $w$.
A random Gaussian matrix $D_x \in \mathbb{R}^{128 \times 256}$ 
is used as a sparsifying dictionary
and $\Phi$ is set equal to the $128 \times 128$ identity matrix.
We use $5\%$ of the generated samples for validation
and $10\%$ for testing.

We design a LeSITA~\eqref{eq:LESITA1} and a LISTA~\eqref{eq:lista2} model 
to learn sparse codes of the target signal.
Different instantiations of both models are realized with different number of layers, i.e., $T = \{3, 5, 7\}$.
Average results are presented in Table~\ref{table: compare_sparse} 
in terms of normalized mean square error (NMSE) in dB.
When the involved signals are similar, i.e., $\rho=\{25, 20\}$, 
LeSITA outperforms LISTA substantially.
The SI has a negative effect in reconstruction 
when the support differs in more than $40\%$ positions.
The results also show that deeper models deliver better accuracy.
Moreover, Table~\ref{table: compare_sparse} includes results for SITA~\eqref{eq:SITA} 
after $T= \{3, 5, 7\}$ iterations, for  $\rho=\{25, 15\}$.
%For $\rho=\{25, 15\}$, we compare LeSITA~\eqref{eq:LESITA1} with the iterative SITA~\eqref{eq:SITA},
%by running $T= \{3, 5, 7\}$ iterations of~\eqref{eq:SITA}.
We also run~\eqref{eq:SITA} with the following stopping criteria: %\footnote{The algorithm stops if one is satisfied.}
maximum number of iterations $T_{\max}=1000$, 
minimum error equal to the error delivered by LeSITA ($T=7$) 
for $\rho=\{25, 20, 15, 10\}$.
The respective average NMSE is $\{-32.35, -29.92, -26.88, -23.92\}$~dB
corresponding to $\{688, 375, 305, 308\}$ iterations (on average).
The comparison shows the computational efficiency of LeSITA against SITA.

A second set of experiments involves real data from 
the EPFL dataset.\footnote{https://ivrl.epfl.ch/supplementary\_material/cvpr11/}
The dataset contains spatially aligned pairs of near-infrared (NIR) and RGB images
grouped in nine categories, e.g., ``urban'' and ``forest''.
Our goal is to reconstruct linearly subsampled NIR images 
(acquired as $y = \Phi x$, $\Phi \in \mathbb{R}^{m \times n}$, $m \ll n$)
with the aid of RGB images.
We convert the available images to grayscale
and extract pairs of $16\times 16$ image patches ($n = 256$),
creating a dataset of $500$K samples. 
One image from each category is reserved for testing.\footnote{
In Table~\ref{table:compare_rec}, an image is identified by a code following 
the category name.}

We design a LeSITA-based reconstruction operator  
with each LeSITA and LISTA encoders comprising $T = 7$ layers,
initialized with weights learned from a LeSITA autoencoder. 
The autoencoder model was initialized with a random Gaussian dictionary $D_x \in \mathbb{R}^{256 \times 512}$ 
and trained using~\eqref{eq:LESITAaeloss} with $\lambda_1 = \lambda_2 = 0.5$.
Besides $\mathcal{L}_{2}^{\text{A}} = \sum_{j=1}^{J}\|\alpha_{(j)}-w_{(j)}\|_1$,
we also experiment with $\mathcal{L}_{2}^{\text{B}} = \sum_{j=1}^{J}\|\alpha_{(j)}\|_1+\|w_{(j)}\|_1$.
For every testing image, we extract the central $256 \times 256$ part 
and divide it into $16 \times 16$ patches with an overlapping stride equal to $4$.
We apply CS with different ratios ($m/n$) to NIR image patches.  

We compare our reconstruction operator with
(i) a LISTA-based~\cite{LeCun_LISTA} reconstruction operator with $T=7$ layers,
(ii) a LAMP-based~\cite{borgerding2017amp} reconstruction operator with $T=7$ layers,
(iii) a deep learning (DL) model proposed in~\cite{nguyen2017deep},
and (iv) a multimodal DL model inspired 
from~\cite{ngiam2011multimodal, ouyang2014multi}; note that~\cite{LeCun_LISTA},~\cite{borgerding2017amp} and~\cite{nguyen2017deep} do not use SI.
The multimodal model consists of two encoding 
and  a single decoding branches.
The target and SI encodings are concatenated to obtain a shared latent representation 
which is received by the decoder to estimate the target signal.
Each encoding branch comprises three ReLU layers of dimension $512$.
The decoding branch comprises one ReLU and one linear layer.
In all experiments, the projection matrix $\Phi \in \mathbb{R}^{m \times 256}$ is
jointly learned with the reconstruction operator.\footnote{The model  in~\cite{nguyen2017deep} learns sparse ternary projections.}
Results presented in Table~\ref{table:compare_rec} in terms of peak signal-to-noise ratio (PSNR) show that
LeSITA trained with $\mathcal{L}_{2}^{\text{A}}$ manages to capture the correlation between the target and the SI signals
and outperforms all the other models.
%============================================%
\section{Conclusions and Future Work}
\label{sec:conclusions}
%----------------------------------------------------------%
We proposed a fast reconstruction operator 
for the recovery of an undersampled signal 
with the aid of SI.
Our framework utilizes a novel deep learning model 
that produces coupled representations of correlated data,
enabling the efficient use of the SI 
in the reconstruction of the target signal.
Following design principles that rely on existing convex optimization methods
allows the theoretical study of the proposed representation and reconstruction models,
using sparse modelling and convex optimization theory.
We will explore this research direction in our future work.
%============================================%
\appendix
\label{sec:prox_proof_LESITA}
The proximal operator for (\ref{eq:probleml1-l1}) has
been defined in \eqref{eq:proxLeSITA} as follows:
\begin{equation*}
\xi_{\mu} (u) = \arg\min_{v}  \big\{ \frac{1}{2} \|v-u\|_2^2 + \mu (\|v\|_1+\|v-w\|_1) \big\}.
\end{equation*}
Let us set
\begin{equation}
\label{proxfun1}
h(v) =   \frac{1}{2} \|v-u\|_2^2 + \mu (\|v\|_1+\|v-w\|_1).
\end{equation}
Considering that the minimization of $h(v)$ is separable, for the $i$-th component of the vectors involved in (\ref{proxfun1}), we obtain 
\begin{equation}
h(v_i) =   \frac{1}{2} |v_i-u_i|^2 + \mu (\|v_i\|_1+\|v_i-w_i\|_1).
\end{equation}
Hereafter, we abuse the notation by omitting the index $i$ and denoting as $v$, $u$, $w$ the $i$-th component of the corresponding vectors.

Let $w\geq 0$. Then we consider the following five cases:
\begin{enumerate}
%------------------------------------------------------------------------%
\item{If $0< w< v$ then %%%%%
\begin{equation}
h(v) =   \frac{1}{2} (v-u)^2 + \mu v +\mu(v-w).
\end{equation}
The partial derivative with respect to $v$ is 
\begin{equation}
\frac{\partial h(v)}{\partial v} =   v -u + 2 \mu.
\end{equation}
$h(v)$ is minimized at $\frac{\partial h(v)}{\partial v} = 0$, that is, $v = u - 2 \mu$.
For $v>w$, we obtain $u > w+2 \mu$. Therefore,
\begin{equation}
\label{prox_ia1}
\xi_{\mu} (u)  =   u - 2 \mu, \quad  u>w + 2 \mu.
\end{equation}
}
%------------------------------------------------------------------------%
\item{
If $0 < v < w$, then  %%%%%
\begin{equation}
h(v) =   \frac{1}{2} (v-u)^2  +\mu v +\mu(-v+w) =   \frac{1}{2} (v-u)^2 + \mu w.
\end{equation}
\begin{equation}
\frac{\partial h(v)}{\partial v} =   v -u.
\end{equation}
\begin{equation}
\frac{\partial h(v)}{\partial v} = 0 \iff v=u.
\end{equation}
For $0 < v < w$, we obtain $0 < u < w$, thus,
\begin{equation}
\label{prox_ib1}
\xi_{\mu} (u)  =  u, \quad 0 < u < w.
\end{equation}
}
%------------------------------------------------------------------------%
%%%%%%
\item{If $v<0$, then %%%%%%
\begin{equation}
h(v) =   \frac{1}{2} (v-u)^2  -\mu v -\mu(v-w).
\end{equation}
\begin{equation}
\frac{\partial h(v)}{\partial v} =   v -u - 2 \mu.
\end{equation}
\begin{equation}
\frac{\partial h(v)}{\partial v} = 0 \iff v=u + 2 \mu.
\end{equation}
For $v<0$, we obtain $u + 2 \mu<0$ or $u < - 2 \mu$, thus,
\begin{equation}
\label{prox_ic1}
\xi_{\mu} (u)  =   u + 2 \mu, \quad u < - 2 \mu.
\end{equation}
}
%------------------------------------------------------------------------%
%%%%%
\item{If $v=0$, then %%%%%%
\begin{align}
\Bigg[\frac{\partial h(v)}{\partial v} \Bigg]_{v=0} 	& =    \Big[v-u + \mu \partial [|v|] - \mu \Big]_{v=0} \nonumber \\
									& = -u + \mu [-1, 1] - \mu \nonumber \\
							    		& = -u + [-2\mu, 0],
\end{align}
where $\partial [\cdot]$ denotes the subgradient. Thus,
\begin{equation}
\frac{\partial h(v)}{\partial v} = 0 \iff u=[- 2 \mu, 0],
\end{equation}
and the proximal operator is given by
\begin{equation}
\label{prox_id1}
\xi_{\mu} (u)  =  0, \quad -2 \mu \leq u \leq 0.
\end{equation}
}
%------------------------------------------------------------------------%
%%%%%%%
\item{If $v=w$, then %%%%%%
\begin{align}
\Bigg[\frac{\partial h(v)}{\partial v} \Bigg]_{v=w}  & =    \Big[v-u + \mu  + \mu \partial [|v-w|]  \Big]_{v=w} \nonumber \\
							    		& = w -u + \mu + [-\mu, \mu] \nonumber \\
							    		& =  - u + [w, w + 2 \mu].
\end{align}
Thus,
\begin{equation}
\frac{\partial h(v)}{\partial v} = 0 \iff u=[w, w + 2 \mu],
\end{equation}
and the proximal operator is given by
\begin{equation}
\label{prox_ie1}
\xi_{\mu} (u)  =  w, \quad w \leq u \leq w + 2 \mu.
\end{equation}
}
%------------------------------------------------------------------------%
\end{enumerate}
Therefore, for $w\geq 0$, (\ref{prox_ia1}), (\ref{prox_ib1}), (\ref{prox_ic1}), (\ref{prox_id1}), and (\ref{prox_ie1}) 
result in:
\begin{equation*}
\xi_{\mu}(u) = 
\left\{\begin{alignedat}{2}
&u+2\mu,	\quad	&&u<-2\mu, \\
&0,			\quad	&-2\mu \leq &u \leq 0, \\
&u,			\quad	&0 <&u<w, \\
&w,			\quad	&w \leq &u \leq w + 2\mu, \\
&u-2\mu	,	\quad	&&u> w+ 2\mu.
\end{alignedat} \right.
\end{equation*}
Similarly, we calculate the proximal operator for $w<0$.

\bibliographystyle{IEEEtran}
\bibliography{l1l1refs}
\end{document}